\documentclass{article} 

\pdfoutput=1

\usepackage{nips15submit_e,times}
\usepackage{hyperref}
\usepackage{url}
\usepackage{graphicx}
\usepackage{geometry}
\usepackage{amsmath}
\usepackage{amssymb}
\usepackage{amsfonts}
\usepackage{amsthm}
\usepackage{subfigure}
\usepackage{algorithm}
\usepackage{algorithmic}
\usepackage{wrapfig}

\theoremstyle{remark}

\newcounter{assumption}
\renewcommand{\theassumption}{A\arabic{assumption}}

\newcommand{\beq}{\begin{equation}}
\newcommand{\eeq}{\end{equation}}
\newcommand{\beqa}{\begin{eqnarray}}
\newcommand{\eeqa}{\end{eqnarray}}
\newcommand{\beqan}{\begin{eqnarray*}}
\newcommand{\eeqan}{\end{eqnarray*}}
\newcommand{\ben}{\begin{eqnarray*}}
\newcommand{\een}{\end{eqnarray*}}

\title{Testing Visual Attention in Dynamic Environments}

\author{
Philip Bachman \\
\small{McGill University} \\
\small{\texttt{phil.bachman@gmail.com}} \\
\And
David Krueger \\
\small{Universit\'{e} de Montr\'{e}al} \\
\small{\texttt{dkrueger@email.com}} \\
\And
Doina Precup \\
\small{McGill University} \\
\small{\texttt{dprecup@cs.mcgill.ca}} \\
}

\nipsfinalcopy 

\begin{document}

\maketitle

\begin{abstract} 
We investigate attention as the active \emph{pursuit of useful information}. This contrasts with attention as a mechanism for the \emph{attenuation of irrelevant information}. We also consider the role of short-term memory, whose use is critical to any model incapable of simultaneously perceiving all information on which its output depends. We present several simple synthetic tasks, which become considerably more interesting when we impose strong constraints on how a model can interact with its input, and on how long it can take to produce its output. We develop a model with a different structure from those seen in previous work, and we train it using stochastic variational inference with a learned proposal distribution.
\end{abstract}

\section{Introduction}
\label{sec:introduction}

One can interpret attention, viewed as a behavioural phenomenon, as a necessary adaptation to intrinsic constraints on perception. For example, if an agent exists in an environment which makes 100 bits of information available per clock tick, but the agent is only capable of observing 10 bits per clock tick, then the agent must be careful about how it directs its perceptual capacity around the environment while capturing what bits it can. The attentiveness of the agent arises not from avoiding noise, but from pursuing signal.


We present several tasks designed to test the capabilities of models which combine visual attention mechanisms and sequential decision making. In spite of their simple structure, these tasks become challenging when we impose strong constraints on how a model can interact with its input, and on how many steps it can take to produce its output. The inputs and outputs in these tasks are either time-varying sequences or multiple presentations of a fixed value. The model constructs its output over a sequence of steps, and at each step it can only perform a single \emph{reading} of its current input through a moveable, low-resolution sensor. To succeed at these tasks a model must use short-term memory for aggregating information across multiple sensor readings, to effectively construct its output and guide future use of its sensor. These tasks extend previous work, e.g.~\cite{Ba2015, Ba2015a, Mnih2014a}, by considering time-varying inputs/outputs, and by putting ``time-pressure'' on output construction (when working with a fixed input).

We develop a model suited to these tasks and train it using stochastic variational inference with a learned proposal distribution.\footnote{One could also think of this training method as Guided Policy Search \cite{Levine2013a} -- see \cite{Bachman2015a} for more on this view.} We empirically show that, given its limited perceptual capacity, our model can perform surprisingly well. The tasks are sufficiently difficult to leave clear room for improvement, particularly in terms of how many times the model must attempt a task before learning a successful policy.

\section{Task and Model Descriptions}
\label{sec:model_and_task}

\begin{figure}
\begin{center}
\includegraphics[scale=0.24]{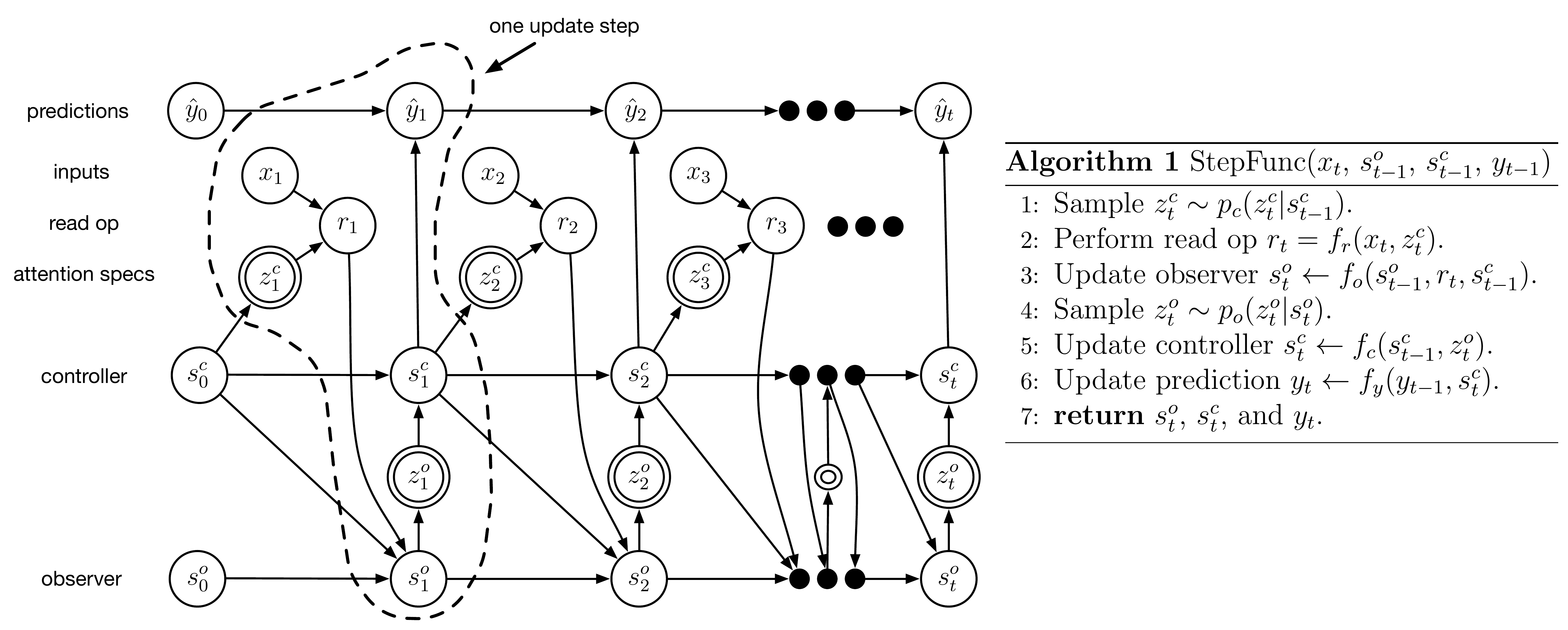}
\vspace{-0.2cm}
\caption{The left figure shows the structure of our model. Single-edged nodes are deterministic and double-edged nodes are stochastic. We don't include the guide module, to keep the figure legible. The guide module parallels the role of the ``observer'', but gets an extra input produced by applying a read op to the gradient of the prediction error. The right figure provides pseudo-code for a single update step in our model.}
\label{fig:seq_lstm_model}
\end{center}
\vspace{-0.5cm}
\end{figure}

\subsection{Our Tasks}

We define tasks based on both static and sequential inputs. In the static input task, which we call ``hurried copying'', the model reconstructs an input using a sequence of readings received from its sensor. At each of several steps, the model decides where to apply its sensor, and how to update its reconstruction of the input. The model thus performs a sort of non-linear adaptive compressive sensing. To make the task more challenging, we limit the number, $T$, and dimension, $k$, of the readings so that the product $kT$ is significantly smaller than the input dimension. We encourage the model to continuously refine its prediction using a cost which simulates evaluating reconstruction at a termination time determined by a poisson random variable.

We also introduce a family of sequential observation and prediction tasks, all of which involve tracking and copying objects from an input video. In these tasks, the model attempts to reconstruct the trajectory of a designated object (or objects) in the presence of noise and/or distractor objects. The model is restricted to observing the inputs through a sequence of low-resolution sensor readings. In our tests we take one reading per video frame. Given a target subset of objects in the input video, the model must reconstruct the input video with all noise and non-target objects removed. Thus, the model must locate and track the target objects while operating under strong constraints on perceptual capacity. Examples are given in Fig.~\ref{fig:videos}.

To generate object paths for our videos, we sample a random (norm-bounded) initial velocity and, at each time-step, we resample a new velocity with probability $.05$. Objects bounce off the image boundary. Object trajectories in our videos are non-deterministic and thus require ongoing prediction and observation for accurate reconstruction. We add uniform noise in $[0,1]$ to each pixel independently with probability $.05$.  Finally, we clip all pixel values to remain within the $[0,1]$ interval.

\subsection{Our Model}

To solve tasks that require interleaving a sequence of perception actions (i.e.~placing and reading from an attention mechanism) with a sequence of updates to a ``belief state'' (which provides the output/prediction at each step), we develop a model built around a pair of LSTMs. We call these LSTMs the ``controller'' and the ``observer''. At step $t$, the observer receives input $r_t$ from an attention module which reads from the current input $x_t$. The read op uses location, scale, precision, etc.~specified by latent variables $z^c_t$ sampled from a distribution conditioned on the state $s^c_{t-1}$ of the controller in the previous step. After the observer updates its own state, its updated state $s^o_t$ determines the distribution of another set of latent variables. A sample $z^o_t$ of these latent variables provides an input to the controller, which updates its own state to get $s^c_t$. The current belief state $\hat{y}_t$ is a function of the previous belief state $\hat{y}_{t-1}$ and the current controller state $s^c_t$. We illustrate this model in Fig.~\ref{fig:seq_lstm_model}.

For the read op, we use a moveable 2x2 grid of differentiable Gaussian filters, as in the DRAW model from \cite{Gregor2015}. We repeat this grid at 1x and 2x scales, for rudimentary foveation\footnote{we omit the 2x scale on our MNIST/TFD experiments.}. At each step we specify the location and scale of the grid, as well as separate ``reading strengths'' (non-negative multipliers) for the 1x and 2x scales. We compute the belief state $\hat{y}_t$ directly as a deterministic function of the controller state $s^c_t$. All functions and distributions in our model depend on trainable parameters. For complete descriptions of our model, our training method, and our train/test data generation, see Github.

To train our model, we add a ``guide module'' which mimics the role of the observer, but receives an additional input produced by applying the read op at time $t$ to the reconstruction residual $y_t - \hat{y}_t$, where $y_t$ indicates the target output at time $t$. The guide module can be interpreted as providing a variational posterior approximation in a directed graphical model, or as the source of guide trajectories in an application of Guided Policy Search. See \cite{Bachman2015a} for more discussion of this view.

\section{Results and Discussion}
\label{sec:empirical_results}

\begin{wrapfigure}{r}{0.4\textwidth}
\vspace{-0.6cm}
\begin{center}
\includegraphics[scale=0.35]{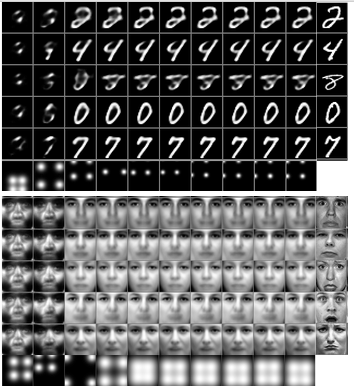}
\vspace{-0.2cm}
\caption{Hurried copying for MNIST and TFD. The right column shows ground truth, and bottom row shows attention.}
\label{fig:hurried_copy}
\end{center}
\end{wrapfigure}

\begin{figure}
\begin{center}
  \subfigure[]{\includegraphics[scale=0.45]{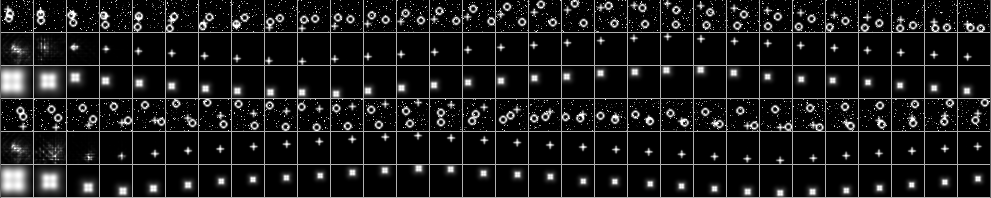}}
  \subfigure[]{\includegraphics[scale=0.45]{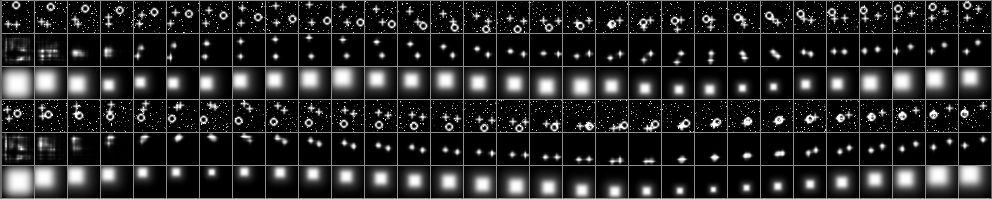}}
  \subfigure[]{\includegraphics[scale=0.45]{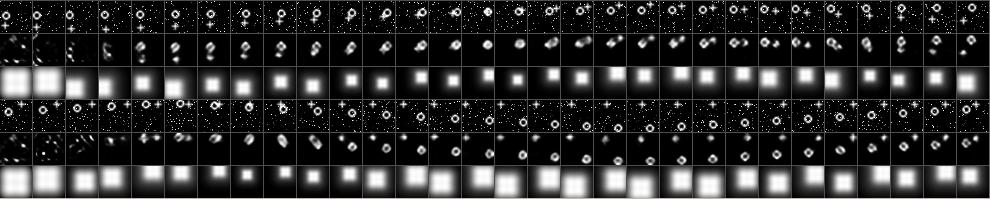}}
  \caption{(a) Model trained with one target (cross) and two distractors (circles). (b) Model trained with two targets (crosses), and one distractor (circle). (c) Model trained with cross and circle targets. In each set of three rows: top is input sequence, middle is reconstruction, and bottom is attention placement.}
  \label{fig:videos}
\end{center}
\vspace{-0.4cm}
\end{figure}

We present results on ``hurried copying'' with TFD~\cite{Susskind2010} and MNIST~\cite{Lecun-98}, and on our detection/tracking tasks with synthetic video data. On image/video tasks, the model pays no reconstruction cost on the first 2/5 frames, respectively. In a sense, this splits each trial into a ``warm-up'' phase and a ``test'' phase. Hurried copy results are in Fig.~\ref{fig:hurried_copy}, and sequential tracking results are in Fig.~\ref{fig:videos}.

Our model reliably learned to track a single object despite the presence of background noise, distractor objects, and random velocity resets. It produced high-fidelity reconstructions of the target object, despite its limited bandwidth sensor. On the more challenging two-object tasks, the model often appeared to follow the objects' mean location while changing the sensor scale and bandwidth to capture both objects in its attention at each step.\footnote{We also observed an interesting mode of behavior in two-object tracking without distractors where the attention mechanism learned to alternate between tracking the two objects in successive time-steps, using distinct foci for each object. The model produced this behavior consistently across different input sequences. Due to time constraints, we have been unable to consistently learn this behavior in two-object tracking with distractors.} While it was able to reconstruct the objects in the correct locations, the reconstructions were not sharp. In general, the cross seemed to be more difficult for the model to reconstruct than the circle. The model also seemed to have more difficulty producing high fidelity reconstructions when tracking objects with different shapes.

On the static image tasks, we observed that the model's first evaluated reconstruction (on the 3rd timestep) is already quite good, despite having only read 12 floating point values at that time. It continues to refine its reconstruction noticeably on MNIST, but these refinements are relatively minor touch-ups. On both datasets, it learned an input-independent attention trajectory. We are investigating the causes of this homogeneity.

\subsection{Discussion}

We presented a set of tasks which provide a test-bed for models combining visual attention with sequential prediction. Though simple in form, our detection and tracking tasks demand sophisticated behavior for consistent success. E.g., to track multiple objects in the presence of distractors, attention must be divided among the target objects while using short-term memory and knowledge of environmental dynamics to estimate the location of objects not currently attended to. We presented a suitable model and some preliminary empirical results showing that our tasks are within reach of current methods, but with plenty of room to grow.


{\tiny
\bibliography{UNIVERSAL-BIB.bib}
\bibliographystyle{plain}
}

\end{document}